\pdfoutput=1
\documentclass[11pt]{article}

\usepackage{ACL2023}

\usepackage{times}
\usepackage{latexsym}

\usepackage[T1]{fontenc}
\usepackage[utf8]{inputenc}

\usepackage{microtype}

\usepackage{inconsolata}

\usepackage[nolist]{acronym}
\usepackage{graphicx}
\usepackage{subcaption}
\usepackage{booktabs}
\usepackage{multirow}
\usepackage{url}
\usepackage{hyperref} 
\usepackage{booktabs}
\usepackage{longtable}
\usepackage{array}
\usepackage{tabularx}

\begin{acronym}
  \acro{ms}[MS]{Multiple Sclerosis}
  \acro{pd}[PD]{Parkinson's Disease}
  \acro{smile}[openSMILE]{Speech \& Music Interpretation by Large-space Extraction}
  \acro{ml}[ML]{Machine Learning}
  \acro{daicwoz}[DAIC-WoZ]{Distress Analysis Interview Corpus from the Wizard-of-Oz interviews}
  \acro{pwms}[pwMS]{people with MS}
  \acro{egemaps}[eGeMAPS]{extended Geneva minimalistic acoustic parameter set}
  \acro{ad}[AD]{Alzheimer's Disease}
  \acro{qol}[QoL]{Quality of Life}
  \acro{ser}[SER]{Speech Emotion Recognition}
  \acro{ai}[AI]{Artificial Intelligence}
  \acro{svm}[SVM]{Support Vector Machine}
  \acro{dl}[DL]{Deep Learning}
  \acro{mfcc}[MFCC]{Mel-Frequency Cepstral Coefficients}
  \acro{cnn}[CNN]{Convolutional Neural Network}
  \acro{uar}[UAR]{Unweighted Average Recall}
  \acro{nn}[NN]{Neural Network}
  \acro{bdi}[BDI-II]{Beck Depression Inventory-II}
  \acro{phq}[PHQ-8]{Patient Health Questionnaire-8}
  \acro{hamd}[HAMD]{Hamilton Rating Scale for Depression}
  \acro{knn}[KNN]{K-Nearest Neighbors}
  \acro{rf}[RF]{Random Forest}
  \acro{cv}[CV]{Cross-Validation}
  \acro{xgb}[XGB]{eXtreme Gradient Boosting}
  \acro{auc}[AUC]{Area Under the Curve}
  \acro{roc}[ROC]{Receiver Operating Characteristic}
  \acro{shap}[SHAP]{SHapley Additive exPlanations}
  \acro{mlp}[MLP]{Multi-Layer Perceptron}
  \acro{llm}[LLM]{Large Language Model}
  \acro{roberta}[RoBERTa]{Robustly Optimised BERT}
  \acro{rnn}[RNN]{Recurrent Neural Network} 
  \acro{vad}[VAD]{Voice Activity Recognition}
  \acro{edss}[EDSS]{Expanded Disability Status Scale}
\end{acronym}

\title{Speech-Based Depressive Mood Detection in the Presence of Multiple Sclerosis: A Cross-Corpus and Cross-Lingual Study}

 \author{Monica Gonzalez-Machorro$^{1,2,3}$, Uwe Reichel$^1$, Pascal Hecker $^{1,4}$, Helly Hammer$^5$,\\ {\bf Hesam Sagha$^1$, Florian Eyben$^{1,6}$, Robert Hoepner$^5$, Björn W. Schuller$^{1,2,3,7}$ } \\ \\ $^1$audEERING GmbH, $^2$TUM University Hospital, \\ $^3$Munich Center for Machine Learning, $^4$Hasso-Plattner Institute, \\ $^5$Inselspital, Bern University Hospital, $^6$Agile Robots, $^7$Imperial College \\ \texttt{monica.gonzalez@tum.de}}

\begin{document}
\maketitle
\begin{abstract}
% HS: possibly use suggestions given by yellow underlines in overleaf
Depression commonly co-occurs with neurodegenerative disorders like Multiple Sclerosis (MS), yet the potential of speech-based Artificial Intelligence for detecting depression in such contexts remains unexplored. 
This study examines the transferability of speech-based depression detection methods to people with MS (pwMS) through cross-corpus and cross-lingual analysis using English data from the general population and German data from pwMS.
Our approach implements supervised machine learning models using: 1) conventional speech and language features commonly used in the field, 2) emotional dimensions derived from a Speech Emotion Recognition (SER) model, and 3) exploratory speech feature analysis.
Despite limited data, our models detect depressive mood in pwMS with moderate generalisability, achieving a 66\% Unweighted Average Recall (UAR) on a binary task. Feature selection further improved performance, boosting UAR to 74\%.
Our findings also highlight the relevant role emotional changes have as an indicator of depressive mood in both the general population and within PwMS. 
This study provides an initial exploration into generalising speech-based depression detection, even in the presence of co-occurring conditions, such as neurodegenerative diseases.
\end{abstract}

\section{Introduction}

\label{sec:intro}

Depression is the most common psychiatric mood disorder \cite{WHO_Depression}. Its prevalence is around 5\%~worldwide \cite{WHO_Depression}.
Despite its prevalence, depression often goes untreated \cite{Johnson2022-helpseeking} due to factors such as socioeconomic barriers and a shortage of healthcare professionals \cite{Evans-socioeconomic}. % HS: I prefer to put refs at the end of the sentence

Speech-based \ac{ai} methods offer a promising approach for fast and non-invasive screening of neurological and mental health during routine examinations 
\cite{speech_new_blood, voice_analysis_for_neurological_disorders}, leveraging speech changes like reduced pitch, slower speaking rate, and articulation errors, which are common in individuals with depression \cite{depression_literature_review}. 
These methods are accessible, scalable, and could enhance help-seeking behaviour and on-going monitoring \cite{Johnson2022-helpseeking}. % HS: 'help seeking behavior' looks a bit strange %MG: this is a common term in pyschology

Prior work has utilised \ac{ml} methods 
to detect depression using acoustic and linguistic features \cite{Kappen2023speech}. 
\citet{mallolragolta19_interspeech}
%Mallol-Ragolta \textit{et al.}(2019)
trained a \ac{rnn} on linguistic features for binary classification on the \ac{daicwoz} dataset, achieving an F1~score of 63\%.
\citet{Zhang2024-depression}~used wav2vec~2.0 for feature extraction and a Long Short-Term Memory (LSTM) network for binary classification using the \ac{daicwoz} dataset, which yielded a 79\%~F1~score. 
% HS: citations at the end of sentence (same style as the next paragraph) %MG: DONE

Similar work has also been conducted in other languages, such as for the German language, \citet{Menne2024thevoice} reported a balanced accuracy 88\% for predicting depressive disorder against healthy controls using acoustic information, and for Italian language, in which \citet{Tao2023theandroid} reported an F1~score of 85\% on the binary task of identifying depression using speech information from a reading task. 

Automatic \ac{ser} research has also been effective in depression detection \cite{wang_ser_importance_2020}, for instance, \citet{wang-depression-ser} developed a \ac{ser} model on the \ac{daicwoz} dataset for binary classification, reporting a 60\%~F1~score. 
% HS: wang-depression-ser is mentioned two times in the same sentence. %mg: CORRECTED

Depression is a common co-morbidity among people with neurodegenerative diseases, such as \ac{ms}, \ac{pd}, and \ac{ad}, among others \cite{depression_comorbidity}, worsening both the \ac{qol} and disease prognosis \cite{hussain2020depression_neurodegenerative}. 
In \ac{ms}, for example,
the lifetime risk of depression is estimated around~50\% \cite{depression_in_ms}.
The overlapping symptomatology of the two conditions can lead to misdiagnosis, with either one of them frequently overlooked \cite{hussain2020depression_neurodegenerative}. 
While prior research highlights the potential of speech-based \ac{ai} methods for depression detection \cite{depression_literature_review}, further work is needed to assess their transferability in patients with neurodegenerative diseases like \ac{ms}.

However, \ac{ms}, due to its impact on the central nervous system, frequently leads to speech impairment, primarily dysarthria \cite{lite_review_ms_dysarthria}. 
As a result, 
\ac{ms} speech typically presents irregular articulatory breakdowns, distorted vowels, pitch breaks, harsh voice quality, and slow speaking rate \cite{lite_review_ms_dysarthria}. 
This raises the question of whether speech-based depression detection can distinguish depressive symptoms in people with a co-existing speech impairment, such as dysarthria, due to a neurodegenerative disease, such as \ac{ms}. %and other disorders affecting speech such as cognitive impairment in \ac{pwms}. 
We hypothesise that these methods would struggle to generalise and distinguish depressive symptoms in \ac{pwms}, since some of the \ac{ms} speech characteristics are similar to those found in people with depression.
% HS: pwms is shown as PwMS in summary, but pwMS here. make them the same %MG: DONE

This contribution aims to address this challenge by assessing the performance of common speech-based methods for depressive mood detection in \ac{pwms}.
To do so, we conduct a cross-corpus and cross-lingual analysis using a well-known English-language 
corpus with depressive mood assessments, along with a German-language dataset of people with low \ac{ms} disability, who also underwent depressive mood assessments.
Our research questions are:
\begin{enumerate}
    \item Do \ac{ml} methods for depressive mood detection generalise to depressive mood detection in \ac{pwms}?
    \item Given that \ac{ser} models have shown promise in detecting emotional changes \cite{wang-depression-ser}, which output from a fine-tuned \ac{ser} model is more effective for depression detection: the model's final results (the classification or regression head output from a \ac{ser} model) corresponding to the emotional dimensions --arousal, valence, and dominance-- or the model's contextualised representations?
    \item Can exploratory feature selection analysis improve generalisability of depression detection in \ac{pwms}?
\end{enumerate}

This contribution is structured as follows. Section \ref{sec:m&m} introduces the datasets, features, and methods employed. Sections \ref{sec:results}, \ref{sec:disc}, \ref{sec:limitation} present the results, limitations, and discussions. Finally, section \ref{sec:conclusion} draws conclusions from the analysis.

\section{Materials and Methods}
\label{sec:m&m}

\subsection{Dataset} 
We 
employ two datasets: 1) The \ac{daicwoz} depression dataset in English presented in \cite{Gratch2014}, and 2) a Swiss German dataset for \ac{pwms} collected under the scope of the COMMITMENT trial \cite{gonzalezmachorro_interspeech_ms}. The trial protocol was approved by national regulatory authorities and local ethic committee (BASEC-ID number 2021-02423) and registered on clinicaltrials.gov (NCT05561621).
The \ac{daicwoz} is a collection of 
semi-structured interviews containing speech samples of 189~participants \cite{Gratch2014}.
It provides predefined speaker-independent training, development, and testing sets, and is segmented at the turn level \cite{valstar_avec2016}. The dataset includes scores from the \ac{phq} self-assessed depression questionnaire.

The COMMITMENT (Prediction of Non-motor Symptoms in Fully Ambulatory MS Patients Using Vocal Biomarkers) 
dataset consists of~50~fully ambulatory \ac{pwms} and 20~control participants. Participants with \ac{ms} have low levels of disability, with a median \ac{edss} score of~$1.0$--indicating minimal impairment-- and a min/max~\ac{edss} score of~$0.0/3.0$, which indicates no disability to moderate disability but still walking unaided. For this paper, we only use the \ac{ms} cohort. Details on the speech recordings are described in \cite{gonzalezmachorro_interspeech_ms}.
Depressive mood scores for each participant are available using the \ac{bdi} questionnaire.
The dataset contains multiple speech tasks. However, in this paper, we utilise two spontaneous speech tasks from each patient: (1) describing the weather on the day of recording and (2) recalling a neutral memory prompted by the word ``grass''. These tasks are chosen because they elicit spontaneous speech and resemble the interview style of the \ac{daicwoz} dataset.
Data was collected using the AISoundLab web platform, which is a
web app, in which each patient could navigate through a voice
recording session under the supervision of a study nurse \cite{gonzalezmachorro_interspeech_ms}. All participants provided informed consent prior to participation, and all data was pseudo-anonymised to protect patient privacy. The ethics consent unfortunately does not permit the publication of the recorded data.

In this paper, participants from the two datasets are categorised as having \textit{depression} or \textit{no depression} based on clinically validated threshold scores from two depression questionaries (\ac{bdi} and \ac{phq}). For the \ac{phq}, participants with a score of~10~or higher are classified as having \textit{depression} \cite{Kroenke2001,Dhingra2011}; and for the \ac{bdi} participants with a score higher than~19 were defined as having \textit{depression} \cite{beckinventory1961}. It is important to keep in mind that these scores serve as indicators of depressive symptoms rather than definitive clinical diagnoses of depression. 

Audio files are downsampled to 16\,kHz. Diarisation for the \ac{daicwoz} data is performed using the turn-level segments provided for each speaker. A \ac{vad} algorithm\footnote{provided by audEERING GmbH} is applied to segment audio files from both datasets, which due to license restrictions, is not open-source. 
For consistency with previous work, we employ the same \ac{vad} parameter values as in \cite{gonzalezmachorro_interspeech_ms}.
Transcripts are automatically obtained for each \ac{vad} segment using Whisper version~2~\cite{whisper_paper} with the \textit{base} model for English and German language.
For the \ac{daicwoz} dataset, we merge the original training and development sets while the original testing set is left intact. The motivation is that due to the small dataset, we opt to use a \ac{cv} strategy for a more robust evaluation. The COMMITMENT dataset, as its purpose is purely for evaluating cross-corpus and cross-lingual generalisation, is not partitioned and it is used as an additional testing set.

Table \ref{tab:participant_per_dataset} describes the metadata for both datasets across the different dataset partitions. Missing values for the questionnaires are dropped before processing. Models trained solely on the COMMITMENT dataset would likely over-fit due to insufficient participants with depressive symptoms to learn acoustic and linguistic markers of depression. Given the imbalance of the two classes, random oversampling with replacement for the two classes and a random seed of~42~is applied. To do so, we employ the package imbalanced-learn \cite{imbalancedlearn}.
%HS: not sure if I understand it! and if you mention random seed, then maybe you need to mention the exact algorithm. %MG: DONE

%%%%%%%%%%%%%%%%%%%%%%%%%%%%%%%%%%%%%%%%%%%%%%%%%%%%%%%%%%%%%%%%%%%%%%%%%%%%%%%

\begin{table}
\centering
\small
\setlength{\tabcolsep}{3.2pt} 
\caption{Metadata for the two datasets employed in this study and the train-test split.}
\label{tab:participant_per_dataset}
\begin{tabular}{p{0.8cm}|p{2.1cm}|p{0.9cm}|p{1cm}|p{1.6cm}}
\toprule
Subset & Dataset & Total Participants & Sex (F/M) & Depression / No Depression \\ 
\midrule
Train & \ac{daicwoz} & 135 & 59 / 76 & 42 / 93 \\ 
\midrule
Test  & \ac{daicwoz} & 44 & 22 / 22 & 13 / 31 \\
       & COMMITMENT & 50 & 37 / 13 & 4 / 46 \\
\bottomrule
\end{tabular}
\end{table}
%%%%%%%%%%%%%%%%%%%%%%%%%%%%%%%%%%%%%%%%%%%%%%%%%%%%%%%%%%%%%%%%%%%%%%%%%%%%%%%%%%%%%%%%%%
\subsection{Feature extraction} We extract six commonly used acoustic and linguistic feature sets, 
and normalise them per dataset using the Robust Scaler, which is robust against outliers. All features are extracted at a \ac{vad} segment-level.
\begin{enumerate}
\item The Wav2Vec2 contextualised representations of length~1024~correspond to the mean pooling of the encoder output. 
These representations are extracted using a publicly available fine-tuned Wav2Vec2 model for 3-dimensional \ac{ser} task \cite{wagner2023dawn}. %\footnote{\url{https://zenodo.org/records/6221127}}.%, and accessible at \url{https://zenodo.org/records/6221127}. 
\item \ac{ser}-dimensions --arousal, valence, and dominance-- are obtained using the same Wav2Vec2 \ac{ser} model \cite{wagner2023dawn}.
These features represent the final outputs of the model returned by the~2-layer multitask regression head \cite{wagner2023dawn}. By extracting both types of information --the contextualised representations and the emotion dimensions-- from the Wav2Vec2 \ac{ser} model, we aim to investigate which one is more effective for depression detection. 
 \item Praat features \cite{Feinberg_2022} are extracted using Nkululeko \cite{Burkhardt:lrec2022} and 
correspond to~39~features, such as voice quality, shimmer, jitter, and duration. This type of features has shown significance for depression detection \cite{depression_literature_review}.
\item \ac{egemaps} \cite{Eyben2016} is extracted using the \ac{smile} feature extraction tool \cite{Eyben2010}. It contains~22~acoustic features related to prosody, voice quality, and articulation. Previous work has reported promising results in depression detection \cite{depression_literature_review}. %, and for \ac{ms} detection \cite{COMMITMENTouscitation}.
We employ the~88~functionals and summary statistics from these features.
\item The psycholinguistic feature set consists of~51~linguistic features that represent the syntactic complexity, the proportion of sentiment tokens, and the proportion of nouns, verbs, negations, adjectives, among others.
\item RoBERTa embeddings are extracted using a multilingual model --XLM Large RoBERTa \cite{conneau2019unsupervised}
--. These embeddings correspond to the $[CLS]$ pooling output applied to the last hidden states of the model. Each segment is defined with a maximum length of~512~tokens and represented by a size of~768.
\end{enumerate}

\subsection{Methods}
We define the following three modelling scenarios to investigate whether \ac{ml} methods for depressive mood detection generalise in the presence of \ac{ms}: 
\begin{description}
    \item[\textbf{A)} Baseline Performance:] Each feature set and model type is trained and evaluated on the \ac{daicwoz} training and testing sets. This task establishes a baseline for model performance in depression detection.

    \item[\textbf{B)} Generalisability Evaluation:] Each feature set and model type is evaluated on the \ac{daicwoz} testing set --to ensure consistent performance on the general population-- and the COMMITMENT dataset. The aim is to assess how well models trained on data from the general population (\ac{daicwoz})
    generalise to the \ac{pwms} data.

    \item[\textbf{C)} Feature Selection Modelling:] Following an exploratory feature analysis on the \ac{daicwoz} training set, the resulting significant features are used for training and evaluation. This task aims to improve model performance by selecting relevant features for depression detection. Two scenarios are investigated:
    \begin{description}
        \item[\textbf{C\_A)}] Models are trained and evaluated on the \ac{daicwoz} training and testing sets using selected features. In other words, it is Task A with selected features. This task assesses whether feature selection improves performance within the general population.
        \item[\textbf{C\_B)}] Models are trained on the \ac{daicwoz} training set using selected features and evaluated on both the \ac{daicwoz} testing set and the COMMITMENT dataset. This scenario, equivalent to Task B with selected features, explores whether feature selection improves generalisability to \ac{pwms} data.
    \end{description}
\end{description}

\textbf{Exploratory feature analysis.}
To investigate which features are significant to distinguish between speakers with and without depression in the training set, we use the Mann-Whitney U test 
($p < 0.05$) because it is non-parametric and does not require the assumption of a normal distribution. This makes it suitable for our data, where not all features follow a normal distribution. Additionally, it is more conservative than other statistical tests, reducing the risk of Type I errors. To quantify the effect size, we use Cohen-R \cite{cohen1988statistical}. Relevant features are found by selecting among the significant ones those with an $r\geq$~.30. Corrections for Type~1~errors are not performed due to the large size of the feature sets, so that the aim of this analysis is restricted to explore acoustic and linguistic feature trends. 

\textbf{Modelling.} We implement supervised \ac{ml} classification for implementing the three modelling tasks. For reproducibility, we seed the pseudo-random number generation. 
The models used are \ac{svm}, \ac{rf}, and \ac{xgb}. These supervised learning algorithms were selected due to their consistently strong performance across a wide range of classification tasks \cite{fernandez2014we}. %Among them, \ac{xgb} has been shown to outperform \ac{rf} in many cases.
Each model is trained using Grid search 5-fold speaker-independent \ac{cv} 
on the training set.

The hyper-parameter values optimised for the Grid Search for each model are as follows: for \ac{svm}, $C \in [10^{-4}, 10^{-3}, 10^{-2}, 10^{-1}, 1, 10]$, the kernel options include \texttt{linear} and \texttt{rbf}, and the gamma parameter is chosen from \texttt{scale} and \texttt{auto}. For \ac{xgb}, the number of estimators $\in [200, 300, 450, 500]$, the learning rate $ \in [0.001, 0.01, 0.1, 0.2]$, the maximum tree depth $\in[4, 5, 6]$, the column subsample ratio $\in[1, 0.3, 0.5]$, and the subsample ratio $\in[0.8, 1]$. Lastly, for the \ac{rf} model, the number of estimators $\in[50, 100, 300, 500, 800, 1000]$, the criterion is either \texttt{gini} or \texttt{entropy}, the minimum number of samples required to split an internal node is $\in[2, 3]$, and bootstrap sampling is either \texttt{True} or \texttt{False}.

The optimal hyper-parameters identified through this process are then used to train the model on the entire training set. 
Class weights are calculated from the training set and are incorporated to address the class imbalance in the data. 

\textbf{Evaluation.} We calculate speaker-level \ac{uar}, F1-score, precision, and recall.
\ac{roc} curves and the \ac{auc} scores were also calculated at a speaker-level. Due to space limitations, only the \ac{roc} curves for the best-performing tasks are presented.
We also
compute the 95\%~Confidence Interval (CI) for the \ac{uar}. The CIs were calculated using 1000~bootstrapping iterations \footnote{\url{https://github.com/luferrer/ConfidenceIntervals}}.

\section{Results}
\label{sec:results}
\begin{table*}[t]
\centering
\caption{Speaker-level test results. \textbf{A}: Baseline Performance. \textbf{B}: Generalisability Evaluation. \textbf{C\_A}: Feature Selection on Task A. \textbf{C\_B}: Feature Selection on Task B. The best-performing combinations for acoustic-based models are marked in \textbf{bold} and *; and linguistic models as \textbf{bold}$^\dagger$. Dep. corresponds to the \textit{depression} class and No Dep. correspond to \textit{no depression}}.
\label{tab:table_results}
\begin{tabularx}{\textwidth}{l l l c c c c c c}
\toprule
Task & Feature & Model & UAR[\%] & F1[\%] & \multicolumn{2}{c}{Precision[\%]} & \multicolumn{2}{c}{Recall[\%]} \\ 
\cmidrule(lr){6-7} \cmidrule(lr){8-9}
& & & & & Dep. & No Dep. & Dep. & No Dep. \\ 
\midrule
\multirow{6}{*}{A} 
    & Wav2Vec2 & \ac{xgb} & 66(49-81) & 65 & 81 & 47 & 71 & 62 \\
    & \ac{ser}-dimensions & \ac{svm} & \textbf{73(57-84)*} & \textbf{67*} & 48 & 90 & 85 & 61 \\
    & Praat & \ac{xgb} & 49(42-62) & 45 & 70 & 25 & 90 & 8 \\
    & \ac{egemaps} & \ac{xgb} & 54(46-69) & 53 & 72 & 50 & 94 & 15 \\
\cmidrule(lr){2-9}
    & Psycholinguistic & \ac{svm} & 46(32-63) & 46 & 68 & 25 & 61 & 31 \\
    & RoBERTa & \ac{svm} & \textbf{56(48-71)$^\dagger$} & \textbf{54$^\dagger$}& 67 & 73 & 15 & 97 \\
\midrule
\multirow{6}{*}{B} 
    & Wav2Vec2 & \ac{svm} & \textbf{66(54-80)*} & \textbf{67*} & 50 & 88 & 41 & 91 \\
    & \ac{ser}-dimensions & \ac{svm} & 64(50-76) & 57 & 28 & 89 & 65 & 64 \\
    & Praat & \ac{svm} & 47(33-60) & 39 & 16 & 79 & 53 & 40 \\
    & \ac{egemaps} & \ac{xgb} & 56(49-69) & 57 & 43 & 84 & 18 & 95 \\
\cmidrule(lr){2-9}
    & Psycholinguistic & \ac{svm} & \textbf{62(48-74)$^\dagger$} & \textbf{54$^\dagger$} & 26 & 88 & 65 & 60 \\
    & RoBERTa & \ac{svm} & 55(49-67) & 54 & 50 & 83 & 12 & 97 \\
\midrule
\multirow{4}{*}{C\_A}  
    & \ac{ser}-dimensions & \ac{xgb} & \textbf{79(70-87)*} & \textbf{70*} & 50 & 100 & 100 & 58 \\
    & Praat & \ac{svm} & 51(36-68) & 51 & 31 & 71 & 31 & 71 \\ 
    & \ac{egemaps} & \ac{xgb} & 58(48-74) & 58 & 60 & 74 & 23 & 94 \\
\cmidrule(lr){2-9}
    & Psycholinguistic & \ac{svm} & 48(33-66) & 48 & 69 & 28 & 58 & 38 \\
\midrule
\multirow{4}{*}{C\_B}  
    & \ac{ser}-dimensions & \ac{xgb} & \textbf{74(60-84)*} & \textbf{65*} & 37 & 93 & 76 & 71 \\ 
    & Praat & \ac{svm} & 46(32-59) & 38 & 16 & 79 & 53 & 39 \\
    & \ac{egemaps} & \ac{xgb} & 57(46-70) & 56 & 29 & 84 & 29 & 84 \\
\cmidrule(lr){2-9}
    & Psycholinguistic & \ac{svm} & 54(42-68) & 51 & 22 & 84 & 41 & 68 \\
\bottomrule
\end{tabularx}
\end{table*}
\subsection{Exploratory feature analysis}
\begin{figure}
    \centering
    \includegraphics[width=0.8\linewidth]{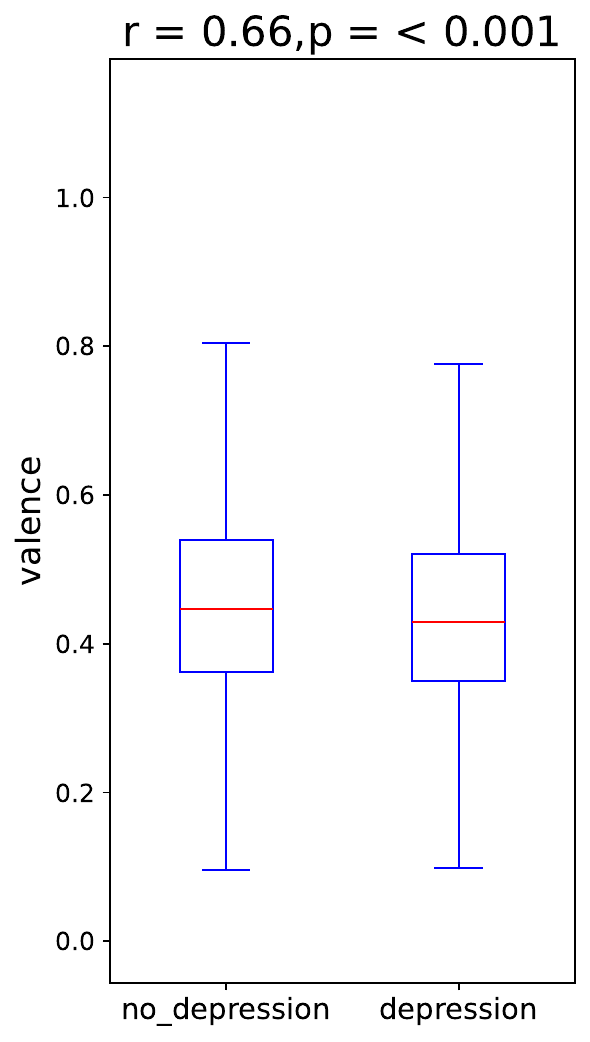}
    \caption{Feature distributions for the binary depression class depression class and valence dimension from \ac{ser} in the \ac{daicwoz} training set. This feature presents a $r$ of 0.66~and~$p<0.001$.}
    \label{fig:correlation_emotions}
\end{figure}

The Mann-Whitney U test is applied to each feature in the training set of the \ac{daicwoz} dataset. Due to interpretability limitations, the Wav2Vec2 and the RoBERTa representations are excluded from the analysis. The number of significant ($p < 0.05$) features with a sufficiently high effect size ($r\geq~0.30$) identified per feature set are: 1) \ac{ser}-dimensions: ~1~feature--valence--; 2) Praat features:~33~out of~39~features; 3) \ac{egemaps}:~64~out of~88~functionals; 4) Psycholinguistic feature set:~18~out of~51~features. These selected features are used in the modelling task C\_A and C\_B to assess whether feature selection improves modelling performance. 
Figure~\ref{fig:correlation_emotions} shows the valence distributions for the binary depression class (``no\_depression'' and ``depression''), which is the only significant features found for the \ac{ser}-dimensions.

\subsection{Modelling Results} 
\begin{figure}
    \centering
    \includegraphics[width=1.0\linewidth]{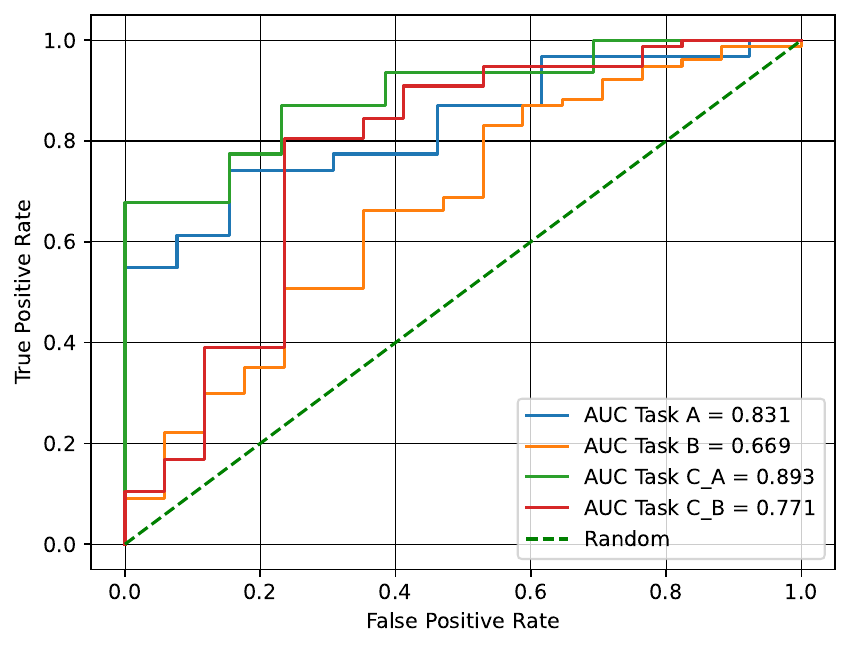}
    \caption{\ac{roc} curve and \ac{auc} value at a speaker-level for the best-performing models using the \ac{ser}-dimenions as feature set across all tasks. Task~A: Baseline Performance. B: Generalisability Evaluation. C\_A: Feature Selection
on Task~A. C\_B: Feature Selection on Task~B.}
    \label{fig:results_auc}
\end{figure}

Table \ref{tab:table_results} shows \ac{uar} and its CIs, F1-score, precision, and recall for \textit{depression} (Dep.) and \textit{no depression} (No Dep.) classes, across the best-performing models and all feature sets. As we are tackling a binary classification
problem, the chance-level \ac{uar} is~50\%.
The best result for Task~A (Baseline Performance) with acoustic features is achieved using \ac{svm} and \ac{ser}-dimensions (\ac{uar}:~73\%), while the best result with linguistic features is achieved using \ac{svm} and RoBERTa embeddings (\ac{uar}:~56\%).
For Task~B (Generalisability Evaluation), Wav2Vec2 embeddings and Psycholinguistic features achieved the best performances (\ac{uar}:~66\% and~62\%, respectively). \ac{ser}-dimensions in Task~B show a performance drop.
For Tasks~C\_A and~C\_B (Feature Selection Modelling on Tasks~A and~B), \ac{xgb} with \ac{ser}-dimensions obtained the highest \ac{uar}s of~79\% and~74\%, respectively.
Since \ac{ser}-dimensions shows consistently good performance in all tasks, Figure~\ref{fig:results_auc} shows the \ac{roc} curves and \ac{auc} values for all tasks. 

\section{Discussion}
\label{sec:disc}

In this paper, we explore three research questions: 
\textbf{1)} Do \ac{ml} methods for depressive mood detection generalise to depressive mood detection in \ac{pwms}? Results in Table~\ref{tab:table_results} indicate that for Task~B (Generalisability Evaluation), acoustic-based features show reasonable generalisability to distinguish depression in \ac{pwms}, with only a modest performance decline compared to results from Task~A (Baseline Performance). 

In the case of the Wav2Vec2 features, a drop in performance for the two tasks is not found, which suggests that these features are transferable to other languages and groups with other co-morbidities such as \ac{ms}. Interestingly, in the case of the \ac{egemaps} features, a minimal increase in performance is observed in Task B, which also suggests a generalisability capacity. 

For Tasks C\_A and C\_B (Feature Selection Modelling), similar patterns are observed as in Tasks A and B, with \ac{ser}-dimensions consistently outperforming other feature sets and demonstrating strong transferability in detecting depression among \ac{pwms}. 
This is further illustrated in Figure~\ref{fig:results_auc}, which highlights the effectiveness of \ac{ser}-dimensions in the context of \ac{ms}.

The top-performing results for Tasks A (using \ac{ser}-dimensions) and B (using Wav2Vec2 features) demonstrate greater precision in predicting the absence of depression (90\% for ``No Dep.'' in Task A; 88\% for ``No Dep.'' in Task B) compared to predicting depression. This finding indicates that identifying depression using speech presents similar challenges in both same-language and cross-lingual contexts, as well as in the general population and among groups with co-morbidities, such as \ac{ms}.

Interestingly, \ac{rf} models did not outperform \ac{xgb} or \ac{svm} in any task or feature set; consequently, they are excluded from Table~\ref{tab:table_results}. This was already reported by \cite{fernandez2014we}, where \ac{xgb} has been shown to outperform \ac{rf} in many cases.

\textbf{2)} Given that \ac{ser} models have shown promise in detecting emotional changes, which output from a fine-tuned \ac{ser} model is
more effective for depression detection: the model's final predictions corresponding to the
emotional dimensions or the model's contextualised representations?
As shown in Table~\ref{tab:table_results}, the \ac{ser}-dimensions and Wav2Vec2 representations achieve the highest \ac{uar} for Task A and Task B, respectively. \ac{ser}-dimensions also outperform all other feature sets in Task C reaching the highest performance. Likely due to the high dimensionality of the Wav2Vec2 embeddings, \ac{ser}-dimensions show overall better results by a small margin. 
However, the performance of \ac{ser}-dimensions and Wav2Vec2 features heavily relies on the performance of the underlying \ac{ser} model \cite{wagner2023dawn}, which was finetuned using the MSP-Podcast dataset (English language) \cite{lotfian2019msppodcast}. It is, therefore, unclear the cross-lingual generalisability of these features when training data would include languages other than English. 

\textbf{3)}~Can feature selection improve generalisability
of depression detection in \ac{pwms}?
Results for acoustic-feature-based models, with the exception of the Praat features, suggest that indeed, feature selection can improve the performance of depression detection. The feature analysis for \ac{ser}-dimensions reveals that only valence among the three dimensions is significantly predictive, highlighting its important role as an indicator of depression in both the general population and \ac{pwms}. This finding is illustrated in Figure~\ref{fig:results_auc}, which shows that individuals without depressive symptoms tend to use higher positive valence in spontaneous interviews compared to those with depressive symptoms. This aligns with prior research, such as
\cite{trifu2024linguisticmarkers}, which found that individuals with depression display lower positive valence than those without.
This pattern may be attributed to a core symptom of depression: emotional dysfunction characterised by a predominant negative emotional state \cite{yang2023emotiondependent}.

\section{Limitations}
\label{sec:limitation}

In the case of text-based models, RoBERTa embeddings achieve above-chance performance in both Task A and Task B while
psycholinguistic-feature-based models exhibit an unexpected trend: their performance on Task B surpassed that of Task A, C\_A, and C\_B. 
The suboptimal performances of text-based models may be due to the use of \ac{vad} segments for feature extraction, which ensured a consistent preprocessing pipeline across acoustic and text features, enabling direct comparisons between model types in detecting depression.
While \ac{vad} segments effectively captured acoustic cues, contributing to strong performances, their short duration may have been less optimal for text-based features, such as word class proportions, which benefit from longer discourse contexts.
The language-specific nature of these features also might have contributed to their struggle to generalise to the German-speaking \ac{ms} population. 
Future work should explore longer segments to optimise text-based models, building on this study’s foundation.

A limitation of this contribution arises from the use of different languages, recording conditions, and depression assessments. Although we try to tackle this by feature normalisation and the restriction to spontaneous speech, further research should explore the impact of language, depression assessments, and recording variations on the generalisability of speech-based depression detection. 
In this paper, we cannot definitively differentiate the extent to which the drop in model performance when evaluated on the \ac{ms} population is influenced by language differences, recording conditions or the presence of \ac{ms} itself.

Moreover, since both \ac{ms} and depression are heterogenous conditions \cite{Gaitan2019ms_misdiagnosis}, implementing personalised approaches when screening for depression in \ac{pwms} is a crucial next step.
Future work should also explore different stages of \ac{ms} --this study focused on low-disability patients-- and account for other co-morbidities in \ac{ms}, like fatigue and cognitive decline, which may also influence speech. 
Also, the \ac{ms} cohort was receiving pharmacological treatment, including common antidepressants for those \ac{ms} patients diagnosed with depression, that could influence mood and, consequently, speech patterns. Although the general population diagnosed with depression from the \ac{daicwoz} dataset may also have been undergoing pharmacological treatments, this information is not available in the dataset, preventing analysis of this potential confounding factor.

To further evaluate the transferability of speech-based depression detection, it is important to examine other common diseases where depression is a common co-morbidity and speech is impacted, such as \ac{pd} or \ac{ad}. A lack of depression scores in speech datasets for these disorders is a major limitation in this regard.
Finally, acoustic and linguistic features alone cannot fully capture the multifaceted nature of depression. These \ac{ml} methods are intended to augment established screening approaches. Incorporating other bio-signals, such as physiological data, could not only enhance performance but also provide a more comprehensive understanding of the disorder.

\section{Conclusion}
\label{sec:conclusion}

In this cross-corpus and cross-lingual study, we explore the efficacy of speech-based depressive mood detection in the presence of \ac{ms} and across English and German languages. 
Our findings highlight the significance of emotional dimensions --arousal, valence, and dominance-- in identifying depressive symptoms, not only in the general population but also within \ac{pwms}. Additionally, acoustic feature sets like \ac{egemaps} also demonstrate potential for generalisability in this context. However, further research is needed to establish robust conclusions.
This study, despite its limitations, represents a step forward towards the integration and generalisability of speech-based depression detection methods. 
Non-invasive speech-based \ac{ai} systems for depression detection hold the potential to improve the \ac{qol} for individuals with this disorder, even in the presence of other illnesses.

\section*{Ethics Statement}
This research was conducted in strict compliance with ethical standards. Data were analysed transparently to minimise bias and ensure robust accuracy. 

% HS: Remove for COMMITMENTity
\section*{Acknowledgements}
We extend our gratitude to all speakers who donated their data. This study was partially funded by Biogen. We thank Biogen's team for coordinating the financial support of the study. 
%We extend our gratitude to all speakers who donated their data. This study was partially funded by Biogen. We thank Biogen's team for coordinating the financial support of the study. 

%\bibliographystyle{acl_natbib}

\end{document}